\documentclass[lettersize,journal]{IEEEtran}
\usepackage{amsmath,amsfonts}
\usepackage{algorithmic}
\usepackage{array}
\usepackage[caption=false,font=normalsize,labelfont=sf,textfont=sf]{subfig}
\usepackage{textcomp}
\usepackage{stfloats}
\usepackage{verbatim}
\usepackage{graphicx}
\usepackage{cite}
\usepackage{url,hyperref,lineno,microtype}
\usepackage[onehalfspacing]{setspace}
\usepackage{times}
\usepackage{todonotes}
\usepackage{epsfig}
\usepackage{amssymb}
\usepackage{cancel}
\usepackage{soul}
\usepackage{outlines}
\usepackage{caption}
\usepackage{booktabs}
\usepackage{multirow}
\def\BibTeX{{\rm B\kern-.05em{\sc i\kern-.025em b}\kern-.08em
    T\kern-.1667em\lower.7ex\hbox{E}\kern-.125emX}}
\usepackage{balance}

\begin{document}

\title{6IMPOSE: Bridging the Reality Gap in 6D Pose Estimation for Robotic Grasping}

\author{Hongpeng Cao\,$^{*}$, Lukas Dirnberger\,$^{ *}$, Daniele Bernardini\,, Cristina Piazza\,, and Marco Caccamo \\ 
    % {\tt\small \{cao.hongpeng, lukas.dirnberger, daniele.bernardini, cristina.piazza, mcaccamo\}@tum.de}
        % <-this % stops a space
\thanks{Hongpeng Cao and Lukas Dirnberger are with School of Engineering and Design, Technical University of Munich, Munich, Germany. {\tt\small \{cao.hongpeng, lukas.dirnberger\}@tum.de}}
\thanks{Daniele Bernardini is with School of Engineering and Design, Technical University of Munich, Munich, Germany and School of Computation, Information and Technology, Technical University of Munich, Munich, Germany. {\tt\small daniele.bernardini@tum.de}} 
\thanks{Cristina Piazza is with School of Computation, Information and Technology, Technical University of Munich, Munich, Germany and Munich Institute of Robotics and Machine Intelligence, Technical University of Munich, Munich, Germany. {\tt\small cristina.piazza@tum.de}}
\thanks{Marco Caccamo is with School of Engineering and Design, Technical University of Munich, Munich, Germany and Munich Institute of Robotics and Machine Intelligence, Technical University of Munich, Munich, Germany. {\tt\small daniele.bernardini@tum.de}}
\thanks{$^{1}$ A video showing the grasping experiments can be found from \url{https://youtu.be/eAc39GzyDIk}; The code, synthetic dataset, and all the pretrained models developed in this work are available at \url{https://github.com/HP-CAO/6IMPOSE}}
\thanks{$^{*}$ These authors have contributed equally to this work and share ﬁrst authorship.}
}
\maketitle

\begin{abstract}
6D pose recognition has been a crucial factor in the success of robotic grasping, and recent deep learning based approaches have achieved remarkable results on benchmarks. However, their generalization capabilities in real-world applications remain unclear. To overcome this gap, we introduce 6IMPOSE, a novel framework for sim-to-real data generation and 6D pose estimation. 6IMPOSE consists of four modules: First, a data generation pipeline that employs the 3D software suite Blender to create synthetic RGBD image datasets with 6D pose annotations. Second, an annotated RGBD dataset of five household objects generated using the proposed pipeline. Third, a real-time two-stage 6D pose estimation approach that integrates the object detector YOLO-V4 and a streamlined, real-time version of the 6D pose estimation algorithm PVN3D optimized for time-sensitive robotics applications. Fourth, a codebase designed to facilitate the integration of the vision system into a robotic grasping experiment. Our approach demonstrates the efficient generation of large amounts of photo-realistic RGBD images and the successful transfer of the trained inference model to robotic grasping experiments, achieving an overall success rate of 87\% in grasping five different household objects from cluttered backgrounds under varying lighting conditions. This is made possible by the fine-tuning of data generation and domain randomization techniques, and the optimization of the inference pipeline, overcoming the generalization and performance shortcomings of the original PVN3D algorithm. Finally, we make the code, synthetic dataset, and all the pretrained models available on Github.
% \footnote{A video showing the grasping experiments can be found from \url{https://youtu.be/eAc39GzyDIk}; The code, synthetic dataset, and all the pretrained models developed in this work are available at \url{https://github.com/HP-CAO/6IMPOSE}.}
\end{abstract}

\begin{IEEEkeywords}
6D pose estimation, RGBD image, Synthetic data, Robotic grasping, Sim2real
\end{IEEEkeywords}

\section{Introduction}
\label{sec:introduction}

Reliable robotic grasping remains a challenge in many precision-demanding robotic applications, such as autonomous assembly \cite{li2019survey} and palletizing \cite{lamon2020towards}. To overcome this challenge, one approach is to accurately recognize the translation and orientation of objects, known as 6D pose, to minimize grasping uncertainty \cite{kleeberger2020survey}. Recent learning-based approaches leverage deep neural networks (DNNs) to predict the 6D object pose from RGB images, achieving promising performance. Nonetheless, estimating 6D poses from RGB images is challenging. Perspective ambiguities, where the appearances of the objects are similar under different viewpoints, hamper effective learning. This problem is further exacerbated by occlusions in cluttered scenarios \cite{zakharov2019dpod}. Additionally, as in many computer vision tasks, the performance of the algorithms is vulnerable to environmental factors, such as lighting changes and cluttered backgrounds \cite{peng2019pvnet}. Furthermore, the use of learning-based methods requires a substantial amount of annotated training data, making it a limiting factor in practical applications as data labeling is time-consuming and costly.

To address the challenges faced by RGB-based approaches, RGBD-based 6D pose estimation algorithms leverage the additional modality from depth images, where the lighting and color-independent geometric information is presented.
One way to leverage depth images is to use the depth for fine pose refinement based on the coarse pose predicted from RGB images \cite{sundermeyer2018implicit, kehl2017ssd}. 
In this case, the initial poses are estimated from the RGB images using DNNs, and the depth information is used to optimize the pose with the Iterative Closest Point algorithm (ICP) to increase the accuracy. 
Another approach is to convert the depth image into point clouds, from which the 6D Pose is predicted \cite{chen2020g2l, hagelskjaer2021bridging}. 
Due to the unstructured nature of the data, working directly on the point cloud is computationally expensive. \cite{chen2020g2l, hagelskjaer2021bridging} first employ an instance detection network to segment the target from the RGB images and crop the point cloud correspondingly. After that, point cloud networks work on the cropped point cloud to predict the 6D pose. 

As an alternative, the geometric features can be directly extracted from the point cloud using DNNs and merged with the RGB features \cite{xu2018pointfusion,li2018unified,wang2019densefusion,he2020pvn3d,he2021ffb6d,lin2022e2ek}. 
Typically, the extracted features of both modalities are matched geometrically and concatenated before further processing \cite{xu2018pointfusion, li2018unified, wang2019densefusion, he2020pvn3d}. This approach is simple to implement and 
%easy for training
simplifies training as the feature extraction networks can be pre-trained in isolation on the available image and point cloud data sets. However, the feature extraction on both modalities could not benefit from each other to enhance representation learning, as the feature extraction DNNs do not communicate. On the other hand, FFB6D \cite{he2021ffb6d} achieves better performance by exploring bidirectional feature fusion at different stages of feature extraction. In this way, the local and global complementary information from both modalities can be used to learn better representations. Moreover, by primarily localizing the target object and excluding the irrelevant background, the feature extraction could be more concentrated on the region of interest, thus the performance can be further improved \cite{lin2022e2ek}.

After feature extraction, different approaches exist to derive the object pose. Direct regression uses dense neural networks to regress to the object's pose directly \cite{yi2018learningorientation}. While this approach allows end-to-end learning and does not require decoding the inferred pose, the optimization of the DNNs is usually difficult due to the limitation of the mathematical representation for the orientation \cite{yi2018learningorientation}. Another common approach is the prediction of orientation-less keypoints and retrieving the pose by their geometric correspondence. \cite{he2020pvn3d, he2021ffb6d, lin2022e2ek} use DNNs to predict the keypoints in 3D space, and then compute the 6D pose via geometry matching on paired predicted keypoints and ground-truth keypoints.

State-of-the-art 6D pose estimation algorithms have achieved excellent performance as evaluated on benchmarks \cite{xu2018pointfusion, wang2019densefusion, chen2020g2l, he2020pvn3d, he2021ffb6d, lin2022e2ek}. However, the training and validation data used in these benchmarks is often correlated, as they are commonly sourced from video frames. Additionally, they may contain environmental features that can bias the learning process and simplify the inference. These factors raise concerns about the generalization of these algorithms and their ability to perform well in real-world scenarios.

Applying the state-of-the-art algorithms to practical robotic applications is non-trivial as the training of 6D pose estimation algorithms has high demand for annotated data \cite{kaskman2019homebreweddb}. 6D pose labeling of images is time and labor intensive, which limits the availability of datasets. On the other hand, using modern simulations to generate synthetic data for training DNNs shows great potential with low cost and high efficiency. For RGB-based approaches, \cite{sundermeyer2018implicit} and \cite{kehl2017ssd} render 3D meshes in OpenGL to generate synthetic RGB images with random backgrounds from commonly used computer vision datasets, for example Pascal VOC \cite{everingham2012pascal} or MS COCO \cite{lin2014microsoft}. Some RGBD approaches \cite{he2020pvn3d,lin2022e2ek,he2021ffb6d,wang2019densefusion} use image composition in RGB and only render depth for the labeled objects. Recently, modern simulations, such as Unity or Blender, enable realistic rendering for full RGBD images, making these engines popular to generate high-quality training datasets \cite{hagelskjaer2021bridging}. 

Unfortunately, the performance of models solely trained on synthetic data often deteriorates when tested on real images due to the so-called \textit{reality-gap} \cite{sundermeyer2018implicit, xiang2018posecnn}. To mitigate the \textit{reality-gap}, domain randomization techniques are often applied to the synthetic data \cite{tobin2017domain}. Domain randomization can be applied to different aspects of image generation. Before rendering, the scene can be randomized by varying the pose of objects, backgrounds, lighting, and the environment to cover as many scenarios as possible \cite{sundermeyer2018implicit, hagelskjaer2021bridging, liu2016ssd}. After rendering, the RGB and depth images can be directly altered, for example changing image contrast, saturation or adding Gaussian blur, and color distortion \cite{sundermeyer2018implicit, kehl2017ssd}. The depth images can be randomized by injecting Gaussian and Perlin Noise \cite{thalhammer2019sydpose} to approximate the noise presented on a real camera. 

Many works \cite{sundermeyer2018implicit, kehl2017ssd, hagelskjaer2021bridging} on 6D pose estimation from synthetic data only evaluate on benchmarks, however, the performance in the real world remains unclear. \cite{zhang_grasping} deploy 6D pose estimation DNNs to real-world robotic grasping, showing promising performance when tested under normal lighting conditions in a structured environment. When tested in unstructured scenarios, where environmental conditions can be inconsistent, the learned algorithms often need real-world data for fine-tuning to bridge the domain gap and achieve comparable performance \cite{li_weakly, deng-self-supervised}. 

The objective of this work is to enhance the reliability of robotic grasping through the use of 6D pose estimation techniques. To address the limitations of existing state-of-the-art approaches and minimize the requirement for manual data labeling, we present 6IMPOSE, a new framework for sim-to-real data generation and 6D pose estimation. The base of 6IMPOSE is the synthetic data generation pipeline that employs the 3D software suite Blender to create synthetic RGBD image datasets with 6D pose annotations. We also include an annotated RGBD dataset of five household objects generated using the proposed pipeline. The object detection module of 6IMPOSE consists of a real-time two-stage 6D pose estimation approach that integrates the object detector YOLO-V4 \cite{bochkovskiy2020yolov4} and a streamlined, real-time version of the PVN3D 6D pose estimation algorithm \cite{he2020pvn3d} for time-critical robotics applications. Furthermore, we provide the codebase to facilitate integration of the vision system into a robotic grasping experiment.

We evaluate the proposed 6IMPOSE framework and 6D pose estimation algorithm on the LineMod dataset \cite{hinterstoisser2012model}. The results show a competitive performance with 83.6\% pose recognition accuracy, comparable to the state-of-the-art methods. It is noteworthy that the models were trained on synthetic data that was uncorrelated to the validation data. To validate the effectiveness of the proposed approach in real-world scenarios, we conducted robotic grasping experiments under varying lighting conditions and achieved an overall success rate of 87\% for five different household objects. To the best of our knowledge, this work is the first to systematically and successfully test sim-to-real 6D pose estimation in robotic grasping. As a contribution to the robotic grasping and related communities, we make the code, synthetic dataset, and all the pretrained models available on Github.

This work is structured as follows: Section \ref{sec:introduction} describes the background of 6D pose estimation in robotic grasping and related work for the proposed data-generation pipeline and two-stage pose estimation algorithm. The design and implementation for the proposed approaches are introduced in details in Section \ref{sec:method}. The proposed approach is evaluated with experimental setup discussed in Section \ref{sec:experiments} and results presented in Section \ref{sec:results}. Section \ref{sec:conclusions} concludes the work and gives a brief outlook on future work.
\section{Methods}

\begin{figure*}[h!]
\begin{center}
\includegraphics[width=\linewidth]{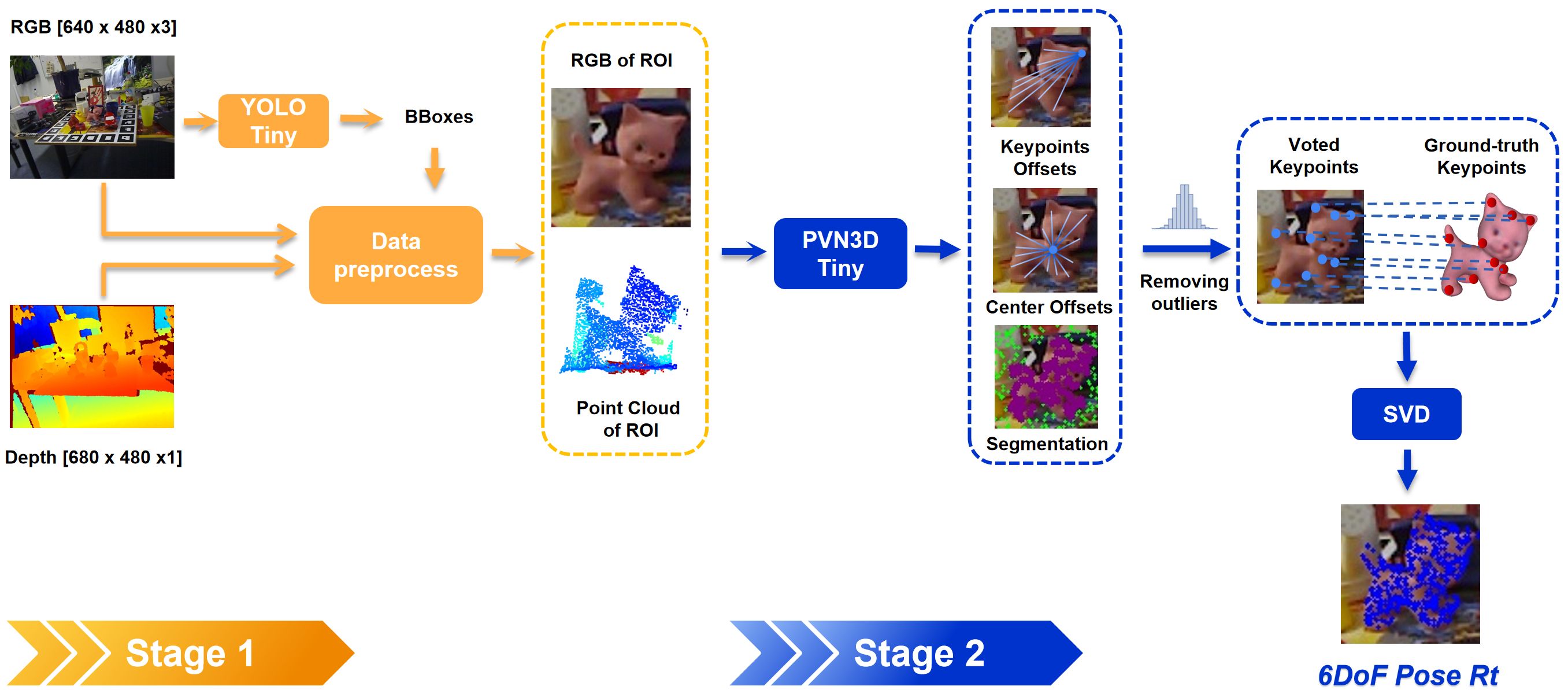}
\end{center}
\caption{A two-stage pose estimation approach showing the object detection with YOLO-tiny to localize the object of interest at the first stage, followed by the 6D object pose estimation with PVN3D-tiny at the second stage.}\label{fig:two-stages}
\end{figure*}

In this section, we first introduce a data preparation pipeline for synthetic data generation and augmentation. Second, we present a two-stage approach to solve the 6D pose estimation problem in real time for robotic applications.

\label{sec:method}
\subsection{Synthetic data generation}
In this work, the synthetic data is generated in Blender \cite{blender} by leveraging its state-of-the-art raycasting rendering functionality. To render RGBD images, a textured 3D model of the object is required, which can be derived from CAD data or collected by 3D scanning.

\textbf{Image Generation}
Given a set of objects, we generate a separate dataset for each object of interest, with the other objects and additional unrelated objects acting as distractors. For each scene to be rendered, we randomly place the objects in the camera's view. In order to avoid overfitting on the color during training, we recolor 25\% of the distracting objects with the dominant color of the main object. Moreover, the distractors' optical properties, such as surface roughness and reflectivity, are varied to further increase the variety of generated images. During simulation, the randomly placed distracting objects can severly occlude the main object, which makes the main object not clearly visible, resulting in invalid training data. To avoid this, we check whether the centroid of the main object is occluded, in which case, we move the occluding objects to the back of the main object. 

We sample images from SUN2012 \cite{xiao2010sun} to use as backgrounds in Blender. Instead of adding the backgrounds to the images after rendering, we follow the idea of image-based lighting, where the backgrounds images are physically rendered as infinite spheres around the scene and emit light. Therefore, the backdrop images affect reflections and lighting conditions in the scene. Furthermore, a random number of point lights are added to the scene with arbitrary power and position. Once the scene is constructed, Blender starts rendering to generate RGB and depth images.

\textbf{Labels}
The segmentation mask can be directly rendered using the object ID feature in Blender. The position and rotation of all objects and the camera is known, and the ground truth transformation matrix $Rt$ can be derived accordingly. The labels for each image are then saved to separate JSON files for each image.

\subsection{Data augmentation}
After rendering the images, we apply several augmentation techniques to mitigate the \textit{reality-gap}. Transformations that would change the object's position in the image would invalidate the ground truth labels, the exception being rotations around the central axis of the imaging sensor. Therefore, an efficient method to increase the number of training data is to rotate each image around the central axis and adjust the labels accordingly. We can efficiently multiply our training data by this method and apply the following techniques separately to each rotated image.

\textbf{RGB data augmentation}
The synthetic RGB images are augmented by randomizing saturation, brightness, hue and contrast, sharpening and blurring. Moreover, we add Gaussian and smooth 2D Perlin noise as in \cite{perlin2002noise} to each color channel to cover different environments and sensors.

\textbf{Depth data augmentation}
The synthetic depth images rendered from simulations are noiseless and almost perfect, which is not the case for images obtained from a real depth camera, where the depth values are often inconsistent and incomplete \cite{mallick2014characterizations}. To approximate inconsistent depth values, we introduce Gaussian noise and Perlin noise to augment synthetic depth images. Similar to \cite{thalhammer2019sydpose}, pixel-level Gaussian noise is added to the synthetic depth images resembling a blurring effect. Smooth Perlin noise has been shown to significantly increase performance when learning from synthetic depth data \cite{thalhammer2019syndd}. We create Perlin noise with random frequency and amplitude and add it directly to the depth channel. The introduced Perlin noise shifts each depth point along the perceived Z-axis, resulting in a warped point cloud, similar to the observed point clouds of real depth cameras. 
In real RGBD images, a misalignment can be observed between depth and RGB images. Similar to \cite{zakharov2018bridging}, we use Perlin noise again to additionally warp the depth image in the image plane. Instead of using a 3D vector field to warp the entire depth image, we restrict warping to the edges of the objects. We apply a Sobel filter to detect the edges and obtain edge masks. We then shift the pixels on the edges using a 2D vector field generated using Perlin noise.

The rendered depth images have no depth information where there is no 3D model, resulting in large empty areas between objects. However, it is also very important to simulate plausible depth values for the background \cite{thalhammer2019syndd}. 

The background depth is based on a randomly tilted plane, to which we add a random Gaussian noise. The noise is sampled on a grid over the image and then interpolated. An additional Gaussian noise is sampled from a second grid and again interpolated. Due to the random and independent choices of grid sizes and interpolation for the two grids, we can achieve a wide variety of depth backgrounds. By adding an appropriate offset, we guarantee that the artificial background is in close proximity to the main object; hence, making object segmentation from the background more difficult. The artificial depth background then replaces empty depth pixels in the original synthetic depth image.

In the real depth images, some regions might miss the depth values and are observed as holes due to strong reflections of the object or other limitations of the depth sensor \cite{mallick2014characterizations}. To simulate the missing depth problem, we first generate a random 2D Perlin noise map, which is converted to a binary masking map based on a threshold. This binary masking map is then used to create missing regions in the synthetic depth image. While this method is not an accurate simulation, we found this approximation, in combination with the other augmentation strategies, useful to improve the accuracy of the neural network.

\subsection{A two-stage 6D pose estimation approach}

The goal of 6D pose estimation is to estimate the homogeneous matrix $Rt \in SE(3)$, which transforms the object from its coordinate system to the camera's coordinate system. This transformation matrix consists of a rotation $R \in SO(3)$ and the translation $t\in \mathbb{R}^3$ of the target object. In this work, we use PVN3D \cite{he2020pvn3d} to infer the homogeneous matrix $Rt$ on the cropped region of interest (ROI) identified by a YOLO-V4-tiny \cite{bochkovskiy2020yolov4} object detector. This two-stage approach is shown in Figure \ref{fig:two-stages}.

The RGB image is processed at the first stage using YOLO-V4-tiny, which provides several candidate bounding boxes and confidence scores. The bounding box with the highest confidence score for a specific object determines the ROI. Given the ROI, the cropped area is the smallest square centered on the ROI and including it, that is a multiple of the PVN3D input size (e.g. 80 x 80, 160 x 160, ...). The square cropped images are then resized to 80 x 80 using nearest neighbor interpolation.

Following PVN3D \cite{he2020pvn3d} and PointNet++ \cite{qi2017pointnet++}, the point cloud is enriched by appending point-wise \textit{R, G, B} values and surface normals. We estimate the surface normal vectors by calculating the depth image's gradients and the pixel-wise normals geometrically as in \cite{holzer2012surfacenormal}. Differently from the original PVN3D \cite{he2020pvn3d} implementation, where the nearest neighbor approach is used to compute the normals from unstructured point clouds, calculating normals from structured depth image is more computationally efficient \cite{nakagawa2015surfacenormal}. This also allows us to use a GPU-based gradient filter in TensorFlow. The resulting point cloud is then randomly subsampled to increase computational efficiency.

In the second stage, PVN3D is used for the pose estimation, with PSPNet \cite{zhao2017pyramid} and PointNet++ \cite{qi2017pointnet++} as backbones to extract RGB and point cloud features separately. The extracted latent features are then fused by DenseFusionNet \cite{wang2019densefusion} at pixel level. Because of the resizing of the cropped RGB image, we map the resized features back to the nearest point in the point cloud. Shared MLPs are then used to regress to the point-wise segmentation and keypoints offsets $\{of_i\} \in \mathbb{R}^3$.

To obtain the final object pose, the point-wise segmentation filters out background points and the keypoint offset are added to the input point cloud to get keypoint candidates. In \cite{he2020pvn3d}, keypoint candidates are clustered by using Mean-Shift clustering for the final voted keypoints $\{\hat{kp_i}\} \in \mathbb{R}^3$. However, the Mean-Shift algorithm works iteratively and this prevents an efficient GPU implementation with deterministic execution time.
To make the keypoint voting temporally deterministic, we first select a fixed amount of point cloud points for each keypoint with the smallest predicted offset. 
Compared to random sampling, this selection method already removes those outliers that show a high offset.
To eliminate any further outliers, we filter out any keypoint candidate whose distance to the mean prediction $\mu$ exceeds the standard deviation $\sigma$, i.e. the offsets $of_i$ will be masked out if $|of_i - \mu| > \sigma $. After removing outliers, we apply global averaging on $\{x, y, z\}$ axis to obtain the voted keypoints $\{\hat{kp_i}\} \in \mathbb{R}^3$. We use singular value decomposition (SVD) to find the \textit{SE(3)} transformation matrix $Rt$ between the predicted keypoints $\{\hat{kp_i}\}$ and the reference model keypoints $\{kp_i\}$.

The prediction accuracy is improved by cropping the image to the ROI, as only the relevant part of the data is processed. With the same number of sampling points, the sampled point cloud from the cropped image is denser, providing PointNet++ with richer geometric information for feature processing, which can also be observed in \cite{lin2022e2ek}. Given the cropped input, we could build the PVN3D with only about 8 millions parameters, which is approximately 15\% of the original implementation \cite{he2020supplementary}. In our test on the LineMOD dataset, the reduced PVN3D performs similarly to the original model. We refer to the reduced PVN3D model as \textbf{PVN3D-tiny}.
\section{Experiments}
\label{sec:experiments}

\begin{figure*}[htbp]
  \centering
  \subfloat[Real data]{
        \includegraphics[width=0.3\linewidth]{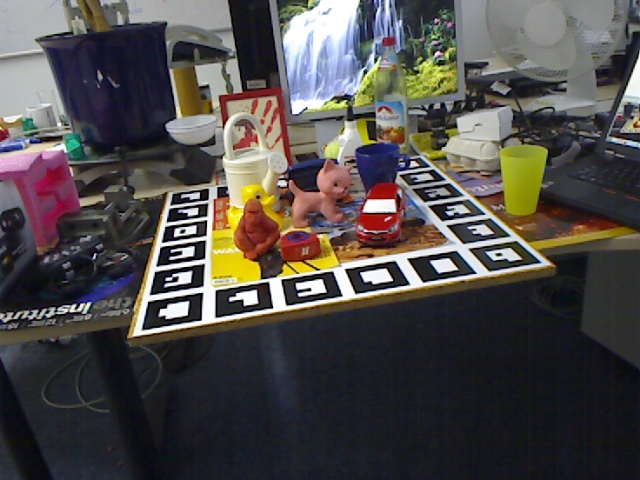}
        \includegraphics[width=0.3\linewidth]{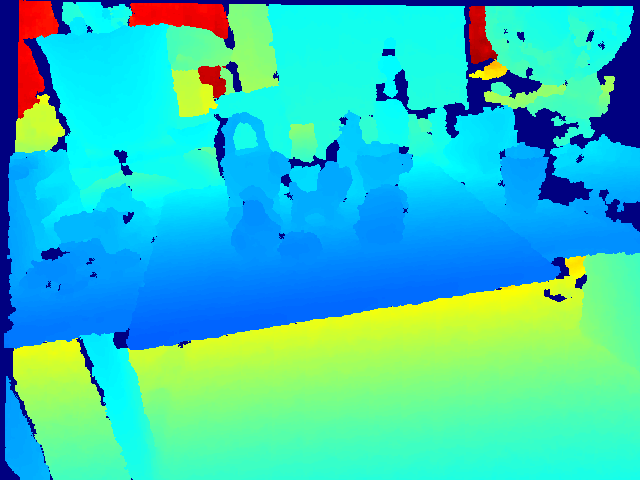} 
        \includegraphics[width=0.3\linewidth]{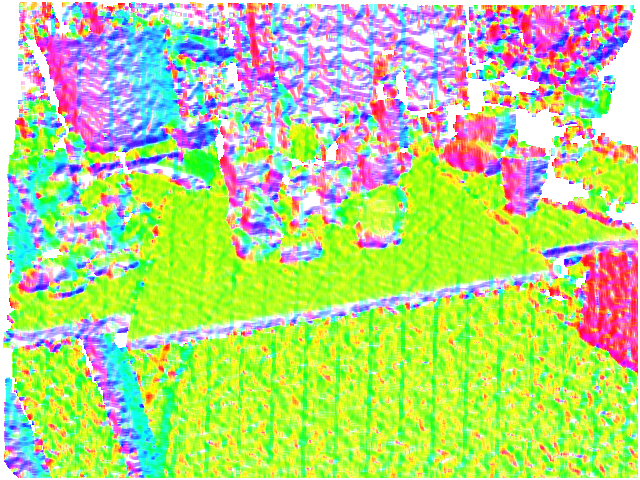}
        \label{fig:real}
  } 
  \hspace{0.05cm}
  \subfloat[Rendered data]{\includegraphics[width=0.3\linewidth]{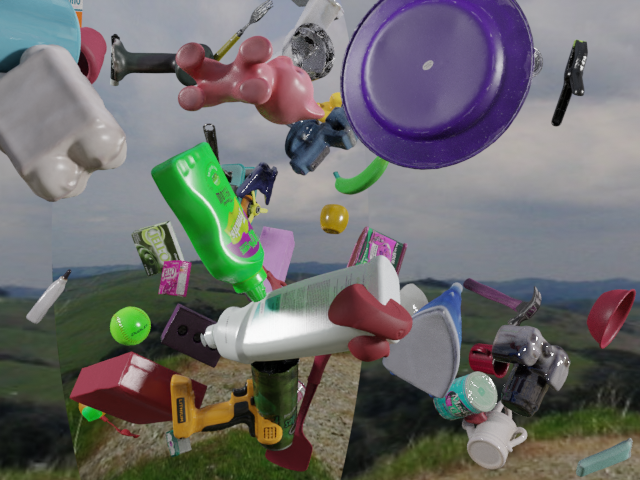} 
        \includegraphics[width=0.3\linewidth]{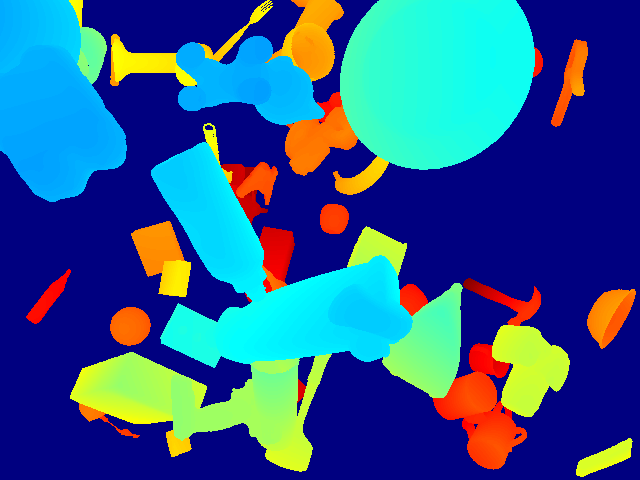} 
        \includegraphics[width=0.3\linewidth]{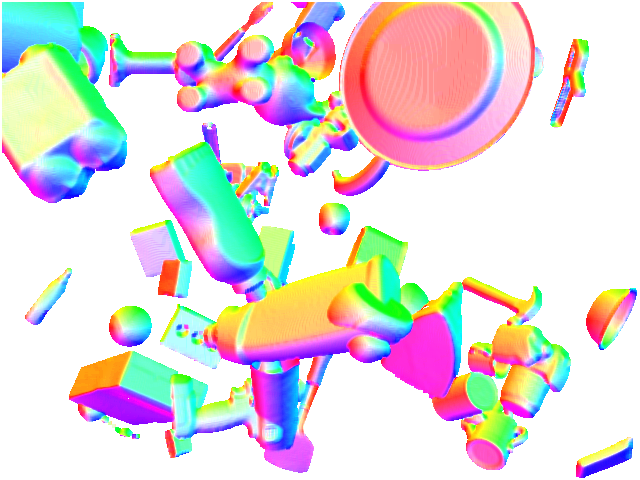}
        \label{fig:render}}
  \hspace{0.05cm}
  
  \subfloat[Augmented data]{\includegraphics[width=0.3\linewidth]{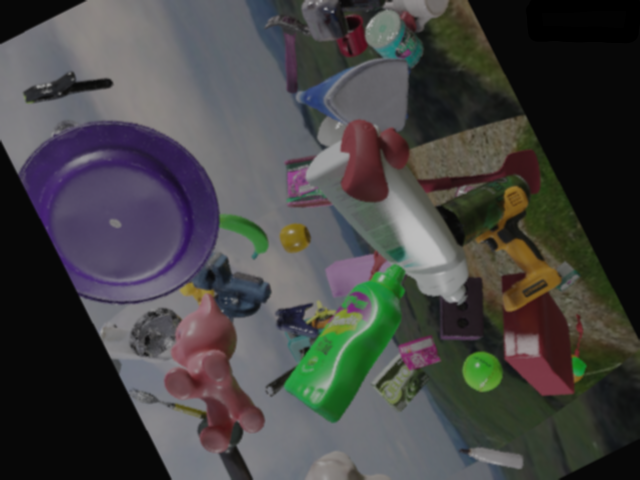} 
        \includegraphics[width=0.3\linewidth]{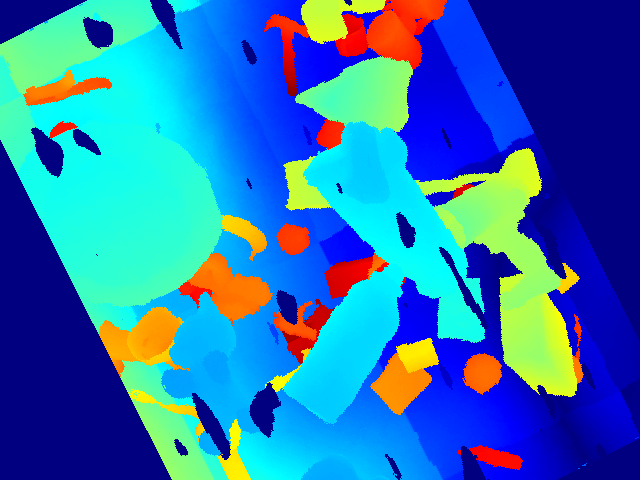} 
        \includegraphics[width=0.3\linewidth]{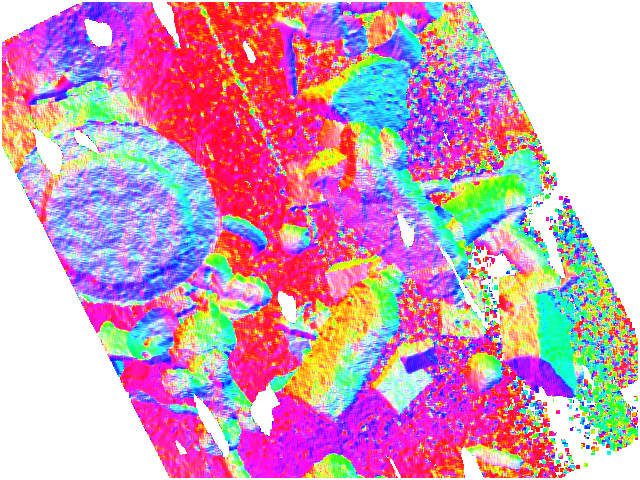} 
        \label{fig:aug}}
  \hspace{0.05cm}
    \caption{The figure showing the visualization of RGB images \textbf{(Left)}, depth images \textbf{(Middle)} and surface normals \textbf{(Right)} for real data \textbf{(a)}, rendered synthetic data \textbf{(b)} and augmented synthetic data \textbf{(c)}.}
    \label{fig:real-synthetic}
\end{figure*}

\begin{figure}[h!]
    \begin{center}
    \includegraphics[width=0.9\linewidth]{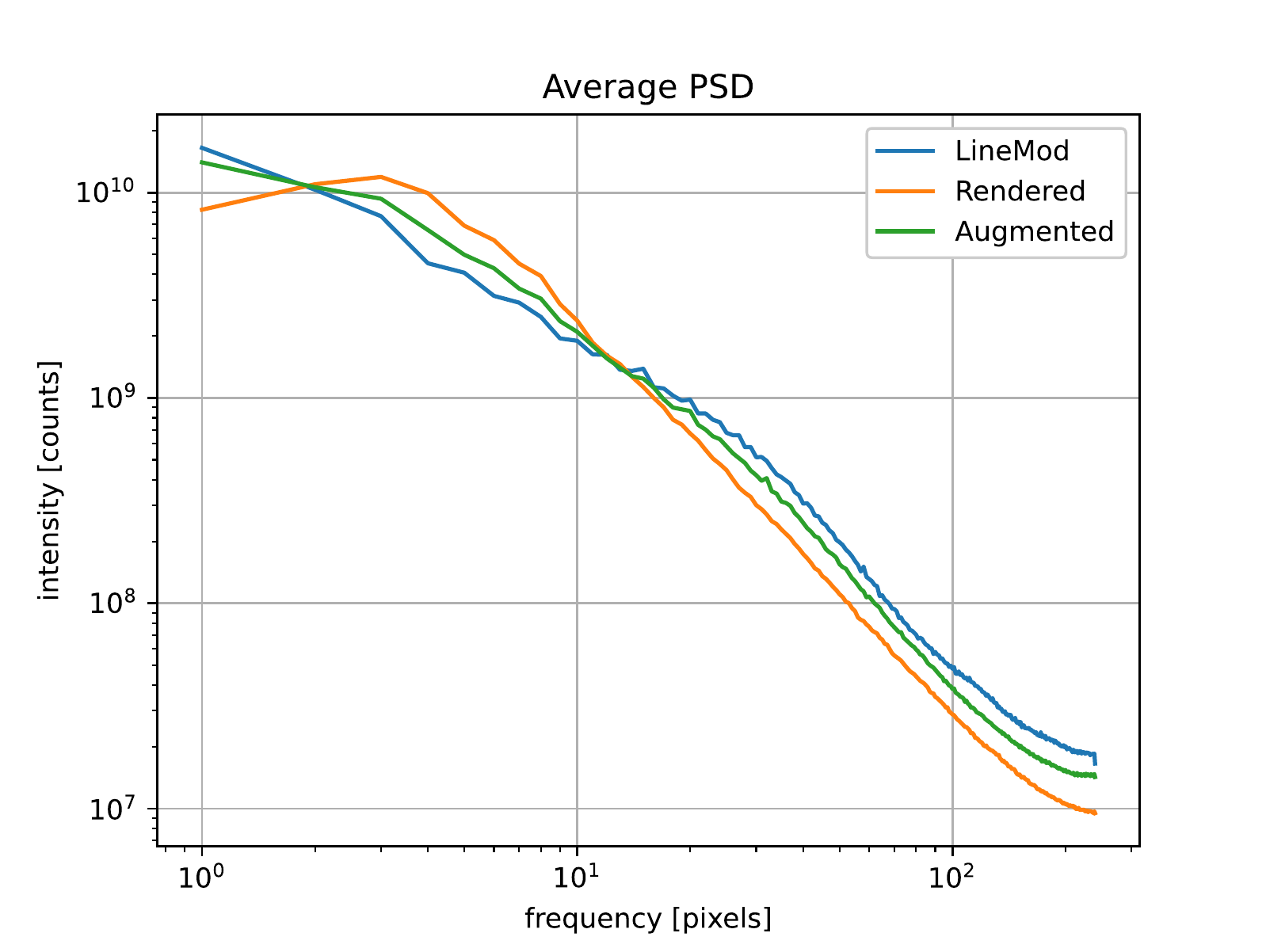}
    \end{center}
    \caption{The plot showing the qualitative average power spectral density (PSD) of depth images with respect to frequencies for the object ``cat" from LM dataset over 50 randomly sampled images.}
    \label{fig:psd}
\end{figure}

\begin{figure}[h!]
    \begin{center}
    \includegraphics[width=0.9\linewidth]{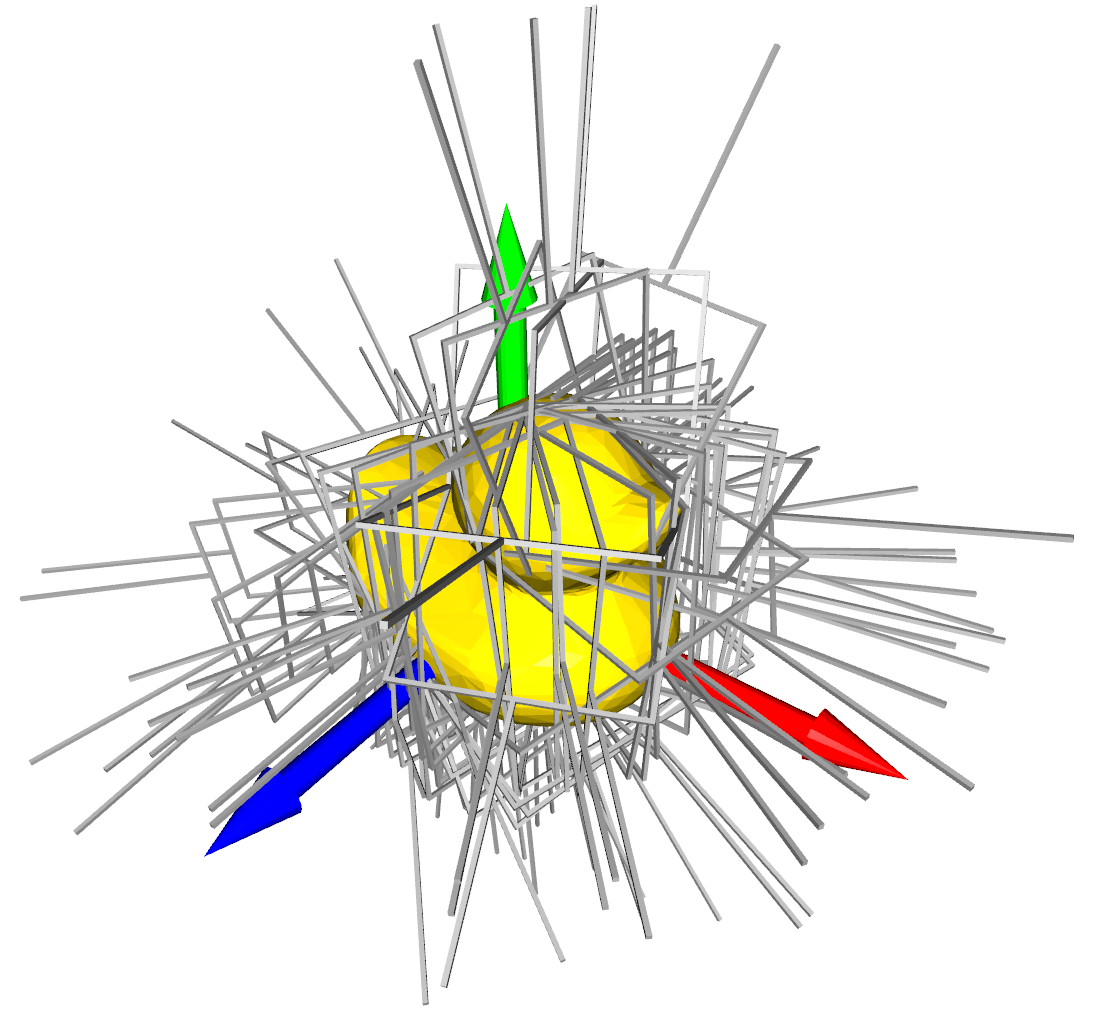}
    \end{center}
    \caption{The figure showing the automatically generated grasp poses for the household object ``duck" .}
    \label{fig:duck_grasps}
\end{figure}

% GRASPING FIGURE

\begin{figure*}[h!]
\begin{center}
    \begin{minipage}[b]{0.9\linewidth}
    \centering
    \includegraphics[width=0.3\linewidth]{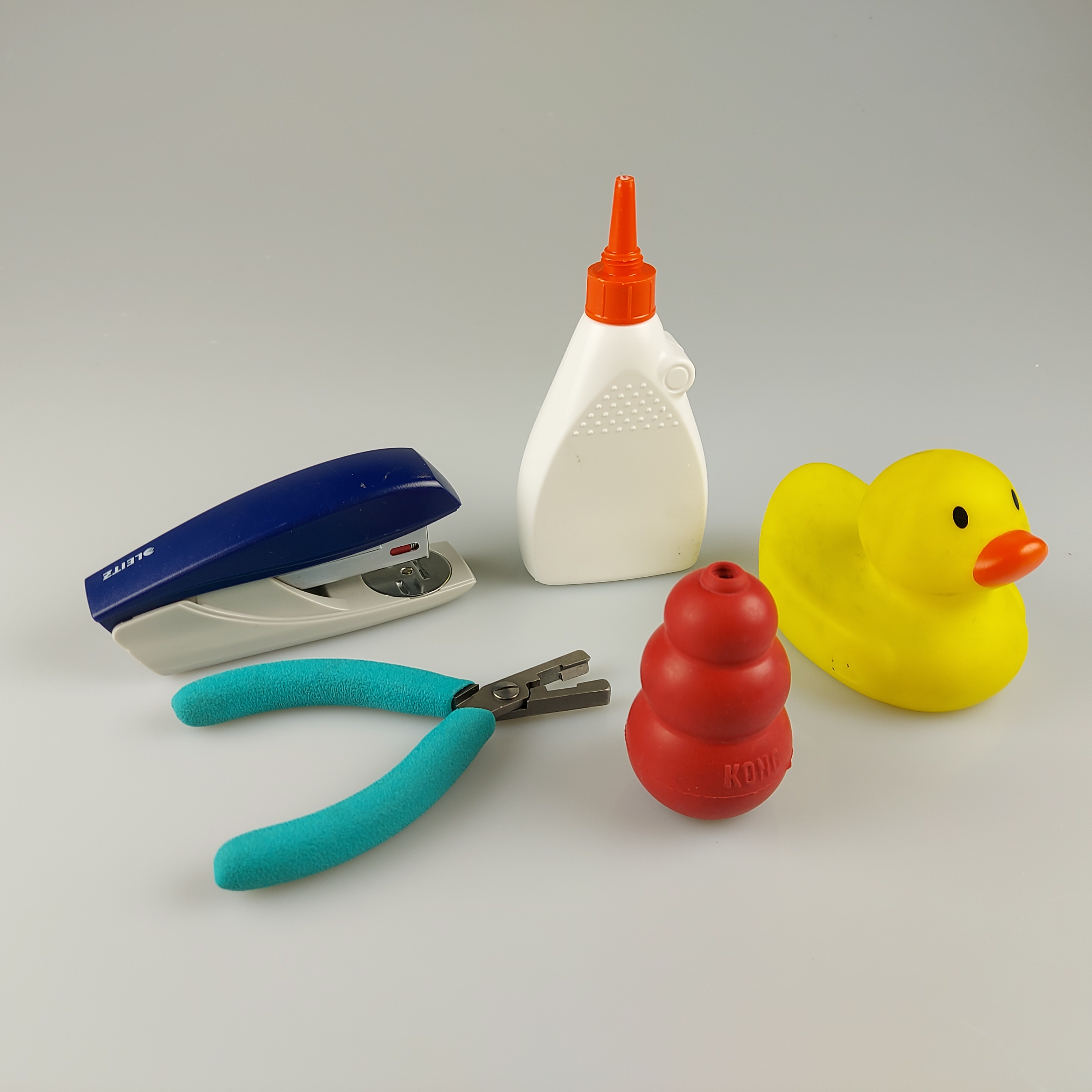}
    \includegraphics[width=0.3\linewidth]{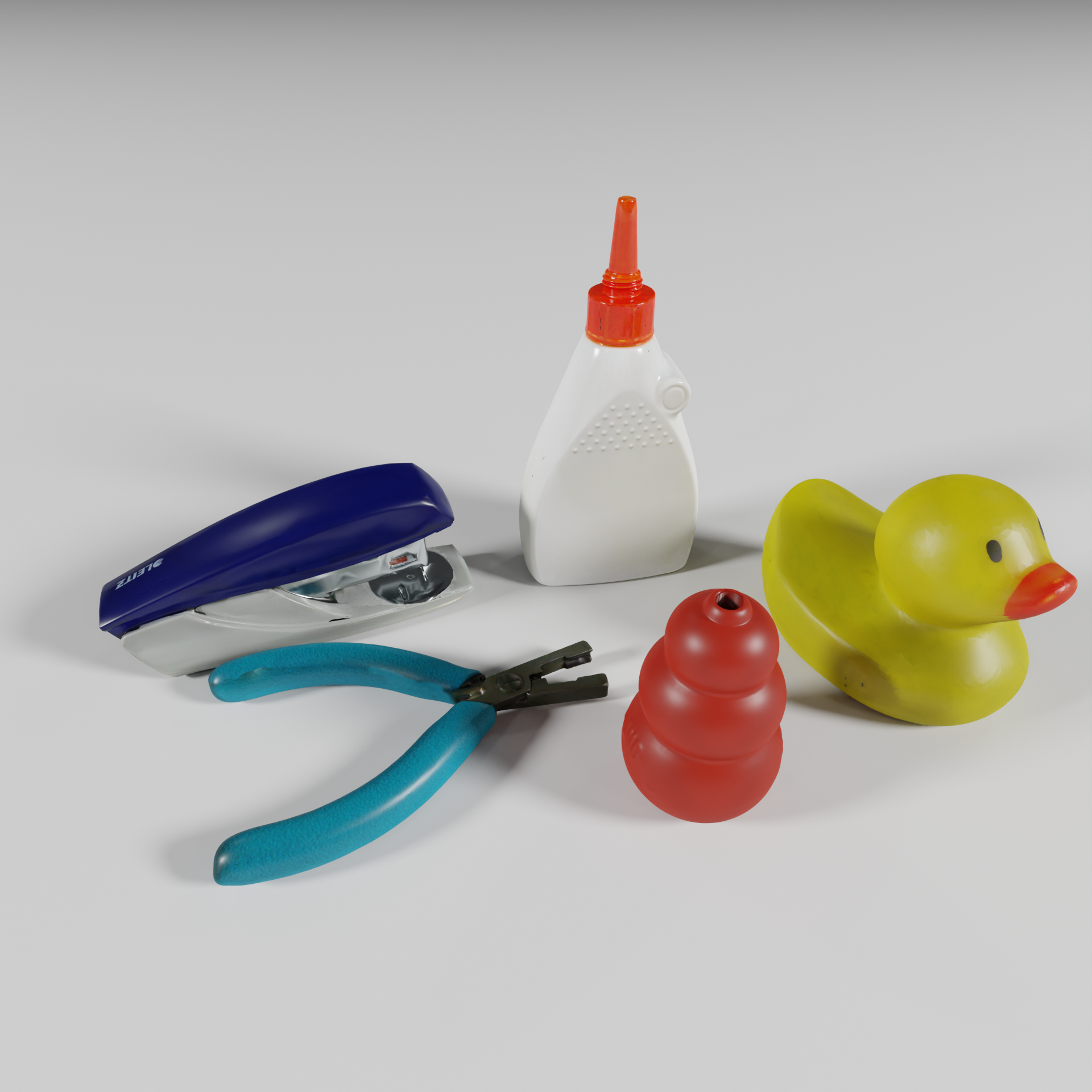} 
    \includegraphics[width=0.3\linewidth]{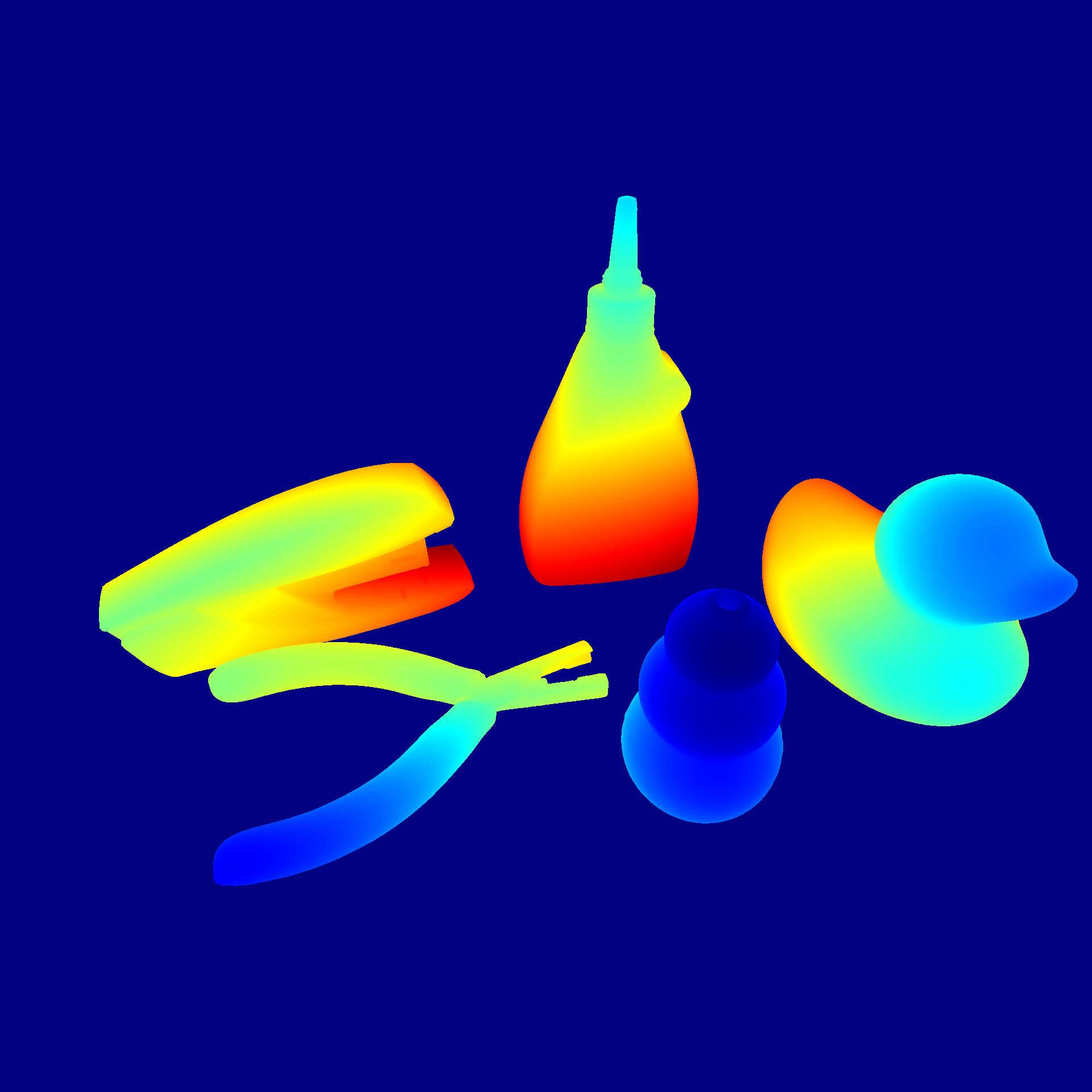}
    \end{minipage} 
    \caption{The figure showing the photographed \textbf{(left)}, photo-realistically rendered RGB \textbf{(middle)} and rendered depth images \textbf{(right)} for the five selected household objects.}
    \label{fig:five_objects}
\end{center}
\end{figure*}

\begin{figure*}
\begin{center}
    \subfloat[Experiments under diffused lighting conditions.]{
    \includegraphics[width=0.24\linewidth]{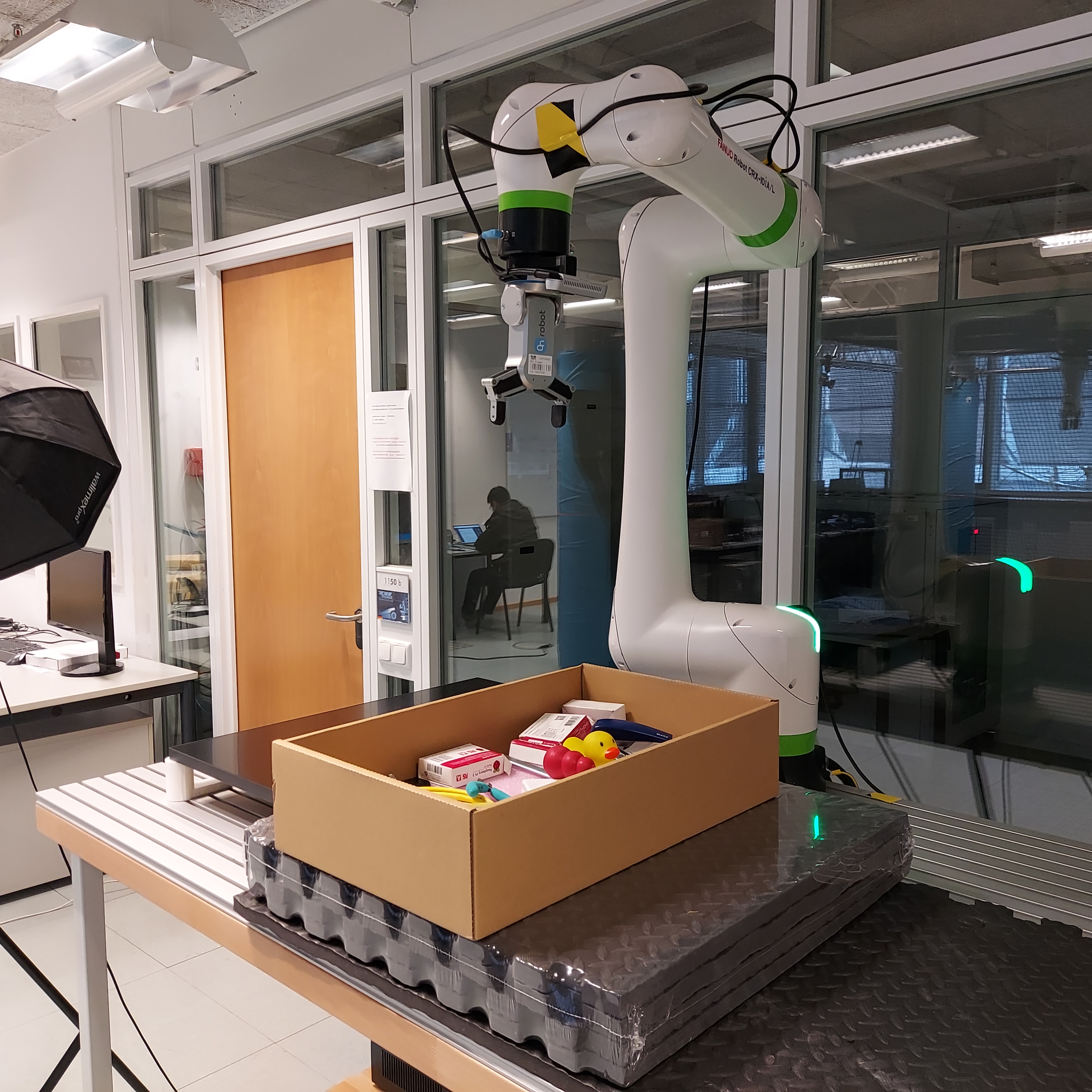}
    \includegraphics[width=0.24\linewidth]{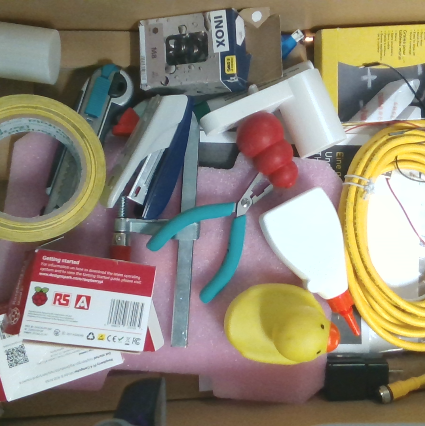} 
    \includegraphics[width=0.24\linewidth]{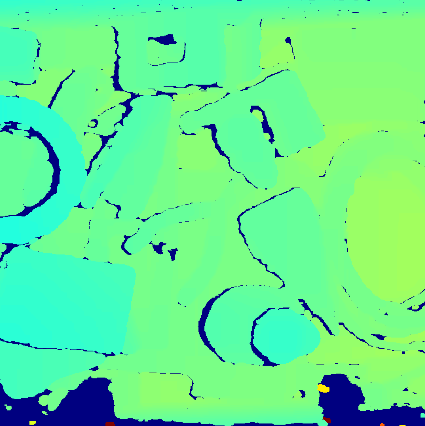}
    \includegraphics[width=0.24\linewidth]{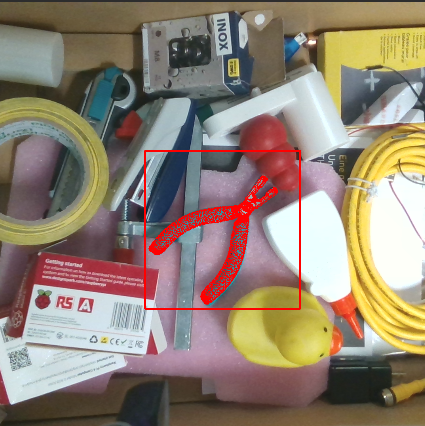}
    \label{fig:exp_normal}}
    \hspace{0.05cm}
    \subfloat[Experiments under low lighting condition.]{
    \includegraphics[width=0.24\linewidth]{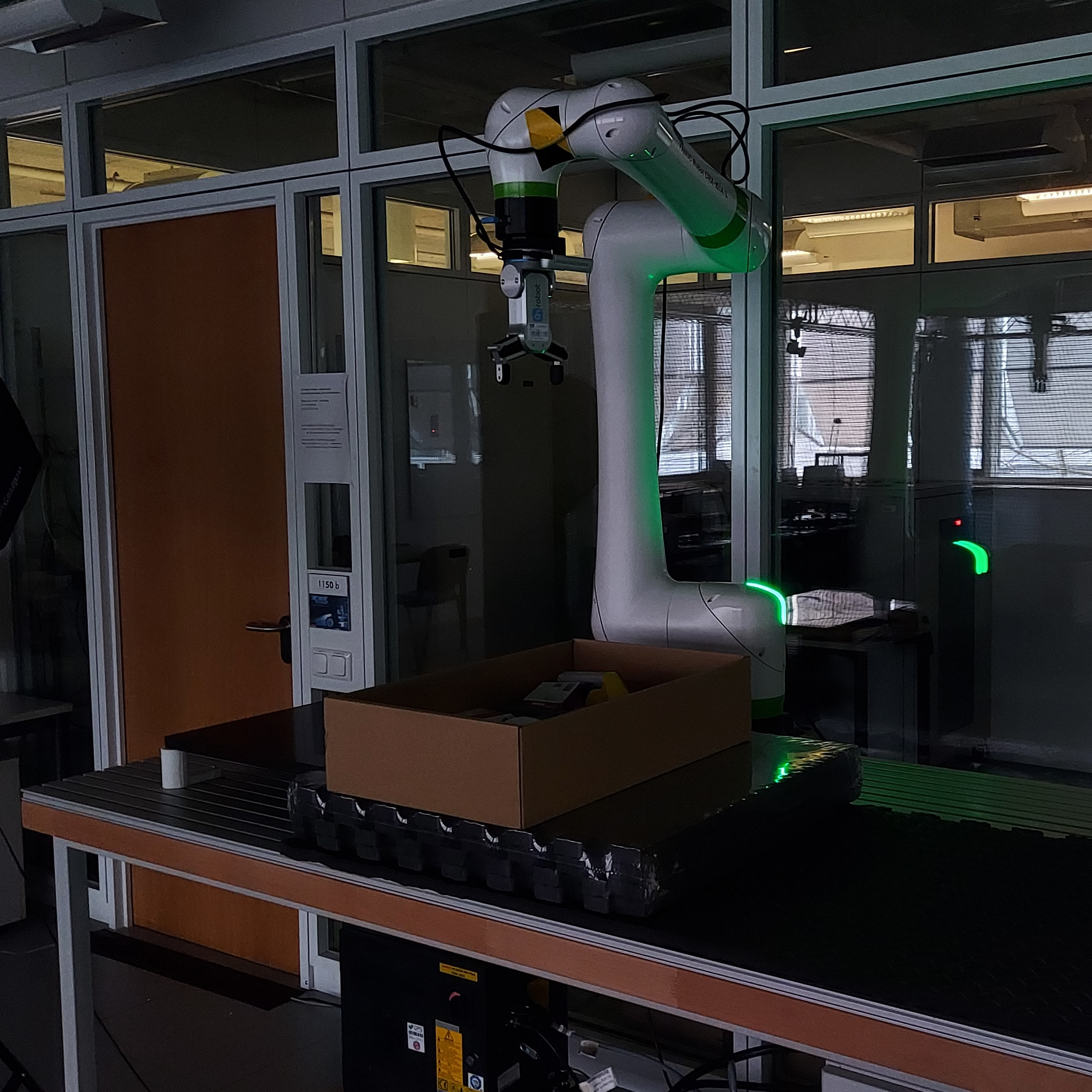}
    \includegraphics[width=0.24\linewidth]{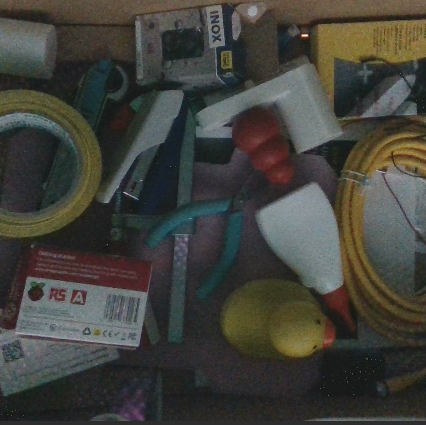} 
    \includegraphics[width=0.24\linewidth]{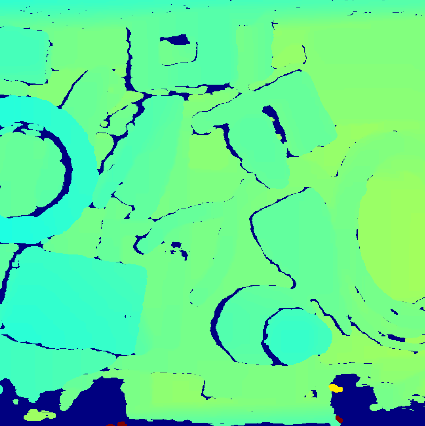}
    \includegraphics[width=0.24\linewidth]{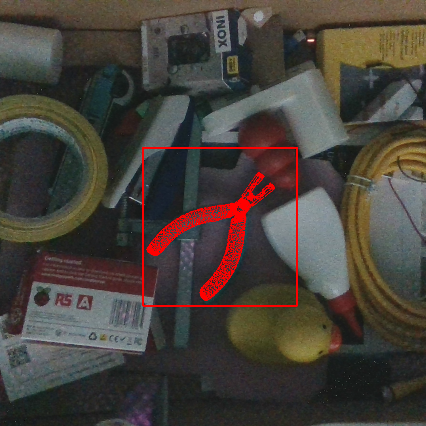}
    \label{fig:exp_low}}
    \hspace{0.05cm}
    \subfloat[Experiments under spot lighting condition.]{
    \includegraphics[width=0.24\linewidth]{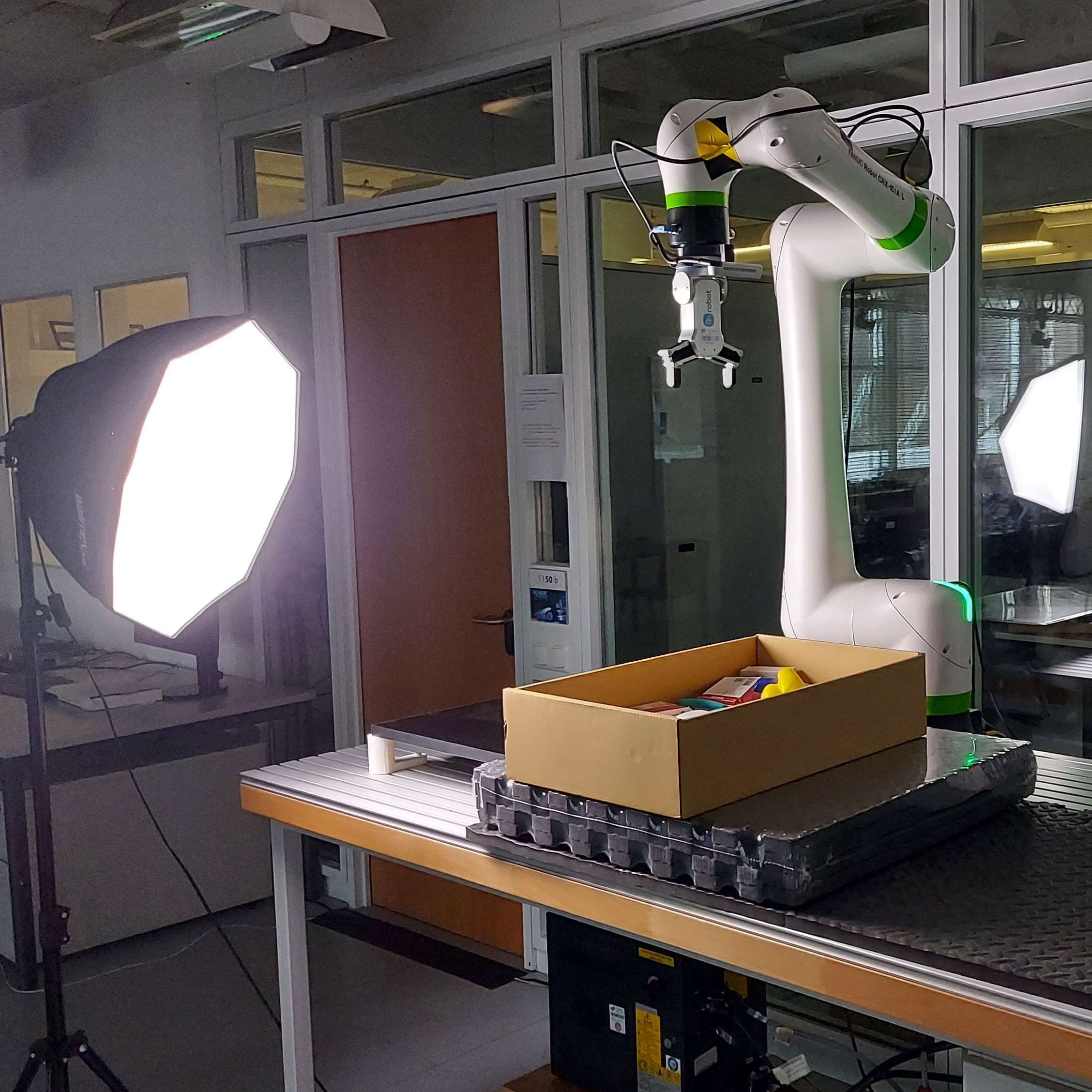}
    \includegraphics[width=0.24\linewidth]{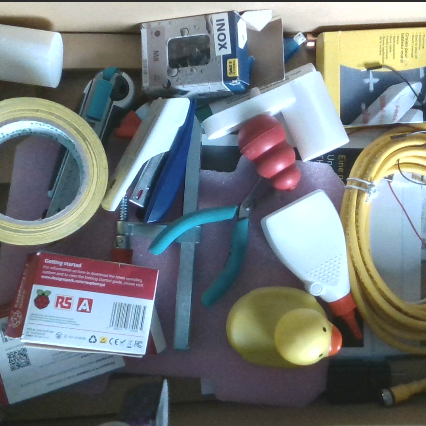} 
    \includegraphics[width=0.24\linewidth]{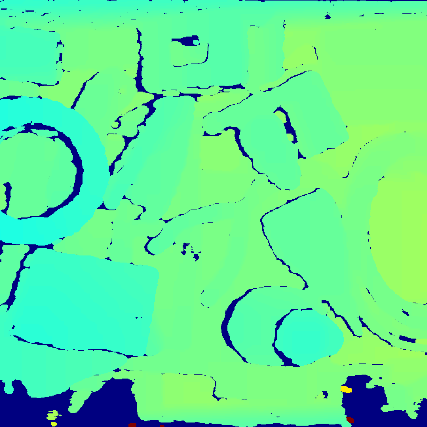}
    \includegraphics[width=0.24\linewidth]{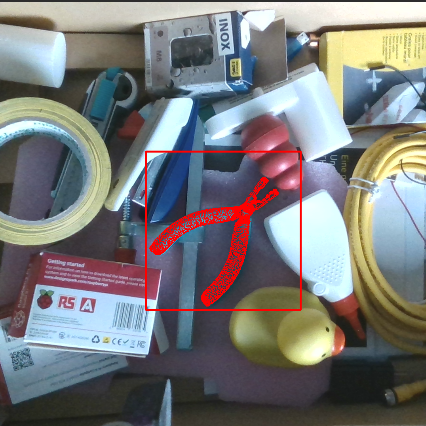}
    \label{fig:exp_spot}}
    \hspace{0.05cm}
    \caption{The figure showing the real world grasping experiments under diffused \textbf{(a)}, low \textbf{(b)}, spot \textbf{(c)} lighting conditions. The experimental setup, RGB view, depth view and the predicted pose of pliers on RGB are shown from left to right.}

\label{fig:real-grasp}
\end{center}
\end{figure*}

In this section, we study the effectiveness of the proposed synthetic data preparation pipeline and the two-stages 6D pose estimation algorithm. Specifically, we use 3D models of objects to generate synthetic RGBD data for training the proposed algorithm. 
After training, we deploy the trained deep models for 6D pose estimation on LinemMod dataset to study ADD(S) pose estimation accuracy and in a real robotic grasping experiment to study grasping success rate (SR).

\subsection{Synthetic data preparation}
In this work, we are interested in the 6D pose estimation of a single object in cluttered environments. For each object, we render scenes in Blender to generate 20k synthetic images using the provided 3D model. We then augment the rendered dataset by rotating each image around the center of the image 16 times, resulting in around 300k images. We discard the images where the object is out of view and pad the images with zeros for empty areas after rotation. Additionally, each image is augmented offline by applying the domain randomization techniques introduced in section \ref{sec:method}. For the training of PVN3D-tiny, we first crop the RGBD to obtain the region of interest according to the ground truth bounding boxes. Furthermore, we generate the point-wise semantic masks and keypoints offsets from the cropped RGBD images using the ground truth poses and segmentation masks.

\subsection{Synthetic data inspection}

To quantify the \textit{reality-gap} between synthetic and real data, we sample 50 RGBD images from the synthetic and real dataset and compare global statistics of these two subsets. For RGB images, we compute the average and the standard deviation for brightness and saturation, as we qualitatively observed that these two factors have a strong influence on the appearances of the generated data. By comparing the statistic of brightness on the synthetic and real subsets, we can optimize the average power and randomization of the point lights and the light-emitting background in Blender. Similarly, with the statistics of saturation, we can optimize the color management in the Blender. To study the statistics of depth images, we use the average power spectral density (PSD) and compare the average distribution on frequencies, as shown in Figure \ref{fig:psd}. Studying PSD on frequencies allows us to inspect the structures of the depth images. And we can accordingly adjust the frequency of the Perlin noise used for depth augmentation. It can be seen that the augmented depth images are closer in frequency distribution to the real images than the non-augmented ones. This is an indication that depth augmentation reduces the gap between the synthetic and real data. 

The examination of global statistics for RGBD images is efficient, as it does not require real annotations. This examination also enables us to identify the "reality gap" qualitatively and adjust the data generation parameters, such as brightness and depth frequencies, to align the synthetic data closer to the real data. Figure \ref{fig:real-synthetic} illustrates examples of real and synthetic images.

\subsection{Implementation}
The synthetic data generation pipeline is implemented in Python using Blender's API. The data randomization and preprocessing are implemented using TensorFlow, accelerating the processing with GPUs. 
As for the two-stage 6D pose estimation approach, we use the original Darknet implementation \cite{bochkovskiy2020yolov4} of YOLO-V4-tiny for the object detection at the first stage and PVN3D-tiny, implemented in TensorFlow, in the second stage.

\subsection{Training and evaluation on LineMod dataset}
To address single object 6D pose estimation problem on LineMod, we separately train a binary YOLO-V4-tiny model and a PVN3D-tiny model for each object of the LineMOD dataset. The YOLO-V4-tiny model is trained using the Darknet framework \cite{darknet13}, and PVN3D-tiny is trained in TensorFlow \cite{abadi2016tensorflow}. All deep neural networks are trained from scratch using only synthetic data without any pretrained models. After training, we build the two-stage 6D pose estimation pipeline by combining YOLO-V4 and PVN3D. We follow \cite{he2020pvn3d} to evaluate the 6D pose estimation performance on the annotated real images provided in LineMOD. The 6D pose estimation performance is measured by using ADD(S) metrics \cite{hinterstoisser2012model}. ADD measures the average distance between the ground truth point cloud and the point cloud transformed with predicted $R,t$, which can be defined as follows: 

\begin{equation}
    ADD = \frac{1}{m}\sum_{v \in \mathcal{O}}{\|(Rv + t) - (R^\ast v + t^\ast)\|}, 
    \label{eq:add}
\end{equation} 

where $m$ is the number of the sampled points, $R^\ast, t^\ast$ is the ground truth pose, and $v \in R^3$ denotes a vertex from the object $\mathcal{O}$. Similarly, the ADDS metric measures the average minimum distance between two point clouds as: 

\begin{equation}
    ADDS = \frac{1}{m}\sum_{v1 \in \mathcal{O}}{\min_{v2 \in \mathcal{O}}{\|(Rv + t) - (R^\ast v + t^\ast)\|}},
    \label{eq:adds}
\end{equation} 

Compared to ADD, ADDS measures the distance to the nearest point instead of correspondent mesh points. For symmetrical objects, ADDS is better suited because ADD yields low scores if the object's pose is different from the ground truth, even if the pose corresponds to an invariant rotation. The success rates on test images are used to quantify the pose estimation performance. A threshold of $10\%$ of the object's diameter is typically used to classify a prediction as successful or not.

\subsection{Robotic Grasping}
We train the proposed approach for pose estimation from purely synthetic images, to perform robotic grasping experiments. We choose five household objects: a rubber duck, a stapler, a chew toy for dogs, a glue bottle and pliers, as shown in Figure \ref{fig:five_objects}, for which the 3D models of the objects are obtained using a Shining3D Transcan C 3D scanner. We generate synthetic training data and train a multi-classes detector YOLO-V4 to localize the target object, and train multi PVN3D-tiny models to estimate poses of different target objects, as described in section \ref{sec:method}. For the grasping experiments, we use a robotic manipulator Fanuc CRX 10iAL with a custom Python interface. As an endeffector, we use an OnRobot RG2 gripper. Attached to the endeffector is a Intel Realsense D415 which is used to obtain the RGBD images. This setup is then used to perform 50 grasp attempts per object in three different lighting conditions, which yields 750 grasps in total. The three different lighting conditions are diffused, low and spot lighting, to test the algorithm's robustness to different lighting levels, as shown in Figure \ref{fig:real-grasp}.

\textbf{Grasping strategy}
The following approach is used to conduct grasping experiments. The robot starts by moving to a predefined home position where the entire bin is visible in the camera's field of view. The object of interest is then identified using YOLO and PVN3D-tiny. To ensure that possible collisions around the object can be observed, the robot moves its end-effector directly above the object. This is important when the object is close to the edge of the camera's view and surrounding obstacles may be out of sight. A safe grasp pose is selected using the pose estimation and grasp selection method. A smooth and tangential trajectory, using a Bézier curve, is generated to approach the object. The gripper is closed when it reaches the grasp pose and the object is lifted out of the bin. To conclude one grasping attempt, the object is dropped back into the bin after the robot returns to the initial home position. If the object can be grasped and lifted without slipping, this grasp will be regarded as a success, a failure otherwise. We distinguish the failure cases between a missed grasp and a collision, to identify the cause of failure, which can be the pose estimation or the collision avoidance.

\textbf{Grasp pose estimation}
Typically, grasp pose estimation follows either an algorithmic or data-driven approach \cite{eppner2021_ACRONYMLargeScaleGrasp, eppner2022_BillionWaysGrasp}. 
In this paper we leverage a simple algorithmic approach similar to \cite{kleeberger2022_AutomaticGraspPose}. Local grasp poses in the object's coordinate frame are generated offline and beforehand. 
With the estimated pose of the object, the local grasp pose can be lifted to the global coordinate frame as a target pose for the robotic manipulator. Generally it is not required to find all grasp poses or the best one, but to find a set of poses, that cover most directions from which the robot may approach the object. Therefore, a list of grasp candidates is generated, that will enable the robotic gripper to securely grasp the object.

We use a sampling based grasp pose estimation using the available mesh of the objects, where randomly sampled points on the surface area of the objects are considered as possible contact points. For each connecting line between a pair of points, we generate 24 grasp poses, rotated around the connecting line and additionally generate the corresponding antipodal grasps. From this grasp candidates, a grasp pose is considered valid if the following criteria are met:
\begin{itemize}
    \item the surface curvature on the mesh should not prevent a stable friction grasp. Therefore, no sharp edges or concave surfaces are considered;
    \item the contact surfaces should be perpendicular to the connecting line. This ensures a stable friction grasp;
    \item the gripper bounding box should not collide with the object. 
\end{itemize}
The remaining grasps are then downsampled, using sparse anchor points in three-dimensional space. 
Our approach typically yields less than 100 grasps for each object, while still providing a high degree of coverage of all possible angles, as can be seen in Figure \ref{fig:duck_grasps}.

\textbf{Grasp pose selection}
The optimal grasp pose for an object of interest is then selected utilizing the predicted 6D pose and the pointcloud data from an RGBD camera.
We first filter out grasp poses that would require the robot to approach the object from a vastly different angle than the current pose of the gripper. 
We then evaluate the remaining grasp poses for collisions by considering every point in the pointcloud not belonging to the object as an obstacle. 
The grasp poses that would lead to a collision with the obstacle points are rejected. Finally, we choose the grasp pose that maximizes the distance to the pointcloud for safety.
\section{Results}
\label{sec:results}

\begin{table*}[htbp]
\caption{The performance of 6D pose estimation on LineMOD compares to the state-of-the-art using RGBD. The bold objects are symmetric.}
\begin{center}
\resizebox{\linewidth}{!}{
{\small{
\begin{tabular}{c|c|c|c|c|c|c|c|c|c|c}
\toprule[1.5pt]
 &\multicolumn{6}{c|}{\textbf{Real data}} &\multicolumn{4}{c}{\textbf{Synthetic Data}}  \\
\cline{2-11} 
 & \textbf{PointFusion}
 & \textbf{DenseFusion$^{\mathrm{*}}$}
 & \textbf{G2L-Net}
 & \textbf{PVN3D}
 & \textbf{FFB6D}
 & \textbf{E2EK}
 & \textbf{AAE$^{\mathrm{*}}$}
 & \textbf{SSD-6D$^{\mathrm{*}}$}
 & \textbf{DGCNN$^{\mathrm{*}}$}
 & \textbf{Ours}  \\

 & \textbf{\cite{xu2018pointfusion}}
 & \textbf{\cite{wang2019densefusion}}
 & \textbf{\cite{chen2020g2l}}
 & \textbf{\cite{he2020pvn3d}}
 & \textbf{\cite{he2021ffb6d}}
 & \textbf{\cite{lin2022e2ek}} 
 & \textbf{\cite{sundermeyer2018implicit}}  
 & \textbf{\cite{kehl2017ssd}} 
 & \textbf{\cite{hagelskjaer2021bridging}} 
 &   \\
\midrule[1.1pt]

Ape  &70.4 &92.3 &96.8 &97.3 &98.4 &98.7 &20.55 & 65 & 97.7 & 78 \\
Benchvise  &80.7 &93.2 &96.1 &99.7 &100 &100 &64.25 & 80 & 99.8 & 92   \\
Camera  &60.8 &94.4 &98.2 &99.6 &99.9 &99.9 &63.20 & 78 & 98.3 & 66 \\
Can  &61.1 &93.1 &98 &99.5 &99.8 &100 &76.09 & 86 & 98.8 & 95 \\
Cat  &79.1 &96.5 &99.2 &99.8 &99.9 &100 &72.01 & 70 & 99.9 & 97 \\
Driller &47.3 &87 &99.8 &99.3 &100 &100 &41.58 & 73 & 99.2 & 91  \\
Duck  &63 &92.3 &97.7 &98.2 &98.4 &99.4 & 32.38 & 66 & 97.8 & 89 \\ 
\textbf{Eggbox} &99.9 &99.8 &100 &99.8 &100 & 100& 98.64 & 100 & 97.7 & 91 \\
\textbf{Glue} &99.3 &100 &100 &100 &100 &100 &96.39 & 100 & 98.9 & 73 \\
Holepuncher  &71.8 &92.1 &99 &99.9 &99.8 &100 & 49.88 &  49 & 94.1 & 61 \\
Iron &83.2 &97 &99.3 &99.7 &99.9 &100 & 63.11 & 78 & 100 & 94  \\
Lamp &62.3 &95.3 &99.5 &99.8 &99.9 &99.9 & 91.69 & 73 & 92.8 & 87  \\
Phone &78.8 &92.8 &98.9 &99.5 &99.7 & 100 & 70.96 & 79 & 99.1 & 74 \\
\midrule[1.1pt]
All &73.7& 94.3 &98.7& 99.4& 99.7& 99.8& 64.67 & 79 & 98.0 & 83.6 \\
\midrule[1.1pt]
\textbf{Speed(s)} & \cancel{}{} & 0.06 & 0.044 & 0.19  &0.075 & 0.068 & \cancel{} & 0.1 &1.0 & 0.046\\
\bottomrule[1.5pt]
\multicolumn{4}{l}{$^{\mathrm{*}}$ With refinement}
\end{tabular}
}}}
\label{tab: LineMOD_evaluation}
\end{center}
\end{table*}

\begin{table*}[htbp]
\caption{6D pose estimation ADD(S) scores, using predicted or ground truth bounding boxes.}
\begin{center}
\resizebox{\linewidth}{!}{
{\small{
\begin{tabular}{c|c|c|c|c|c|c|c|c|c|c|c|c|c|c}
\toprule[1.5pt]
 & Ape & Benchvise & Camera & Can & Cat & Driller & Duck & \textbf{Eggbox} & \textbf{Glue} & Holepuncher & Iron & Lamp & Phone & All \\
\midrule[1.1pt]
Predicted bboxes & 78 & 92 & 66 & 95 & 97 & 91 & 89 & 91 & 73 & 61 & 94 & 87 & 74 & 83.6 \\
GT bboxes & 81 & 99 & 98 & 96 & 97 & 99 & 94 & 99 & 99 & 78 & 96 & 94 & 96 & 94.3 \\
\bottomrule[1.5pt]
\end{tabular}
}}}
\label{tab:LineMOD_ablat}
\end{center}
\end{table*}

\begin{table}[htbp]
\caption{Running time analysis of the proposed two-stage pose estimation approach.}
\begin{center}
\resizebox{0.9\columnwidth}{!}{
{\small{
\begin{tabular}{c|c|c}
\toprule[1.5pt]
Procedures & Speed Mean/Std(ms) & Percent\\
\midrule[1.1pt]
YOLO-V4 tiny & 6.7/0.5 & 15\% \\
Pcld preproc. & 8.2/5.9 & 18\% \\
PVN3D tiny & 23.7/0.9 & 51\% \\
Pose regression  & 7.2/0.7 & 16\% \\
\midrule[1.1pt]
All & 45.8/6.2 & 100\% \\
\bottomrule[1.5pt]
\end{tabular}
}}}
\label{time_analysis}
\end{center}
\end{table}

\begin{table*}[htbp]
\caption{Single Object grasping experiments under varying lighting. conditions}
\begin{center}
\resizebox{0.8\linewidth}{!}{
{\small{
\begin{tabular}{c|ccccccc}
\toprule[1.5pt]
Conditions                                         & \multicolumn{1}{c|}{Category}     & \multicolumn{1}{c|}{Duck}   & \multicolumn{1}{c|}{Stapler} & \multicolumn{1}{c|}{Glue}   & \multicolumn{1}{c|}{Chewtoy$^{\mathrm{*}}$} & \multicolumn{1}{c|}{Pliers} & \textbf{All}                        \\ 
\midrule[1.1pt]
\multirow{5}{*}{Diffused}                          & \multicolumn{1}{c|}{Success}      & \multicolumn{1}{c|}{47}     & \multicolumn{1}{c|}{46}      & \multicolumn{1}{c|}{46}     & \multicolumn{1}{c|}{45}      & \multicolumn{1}{c|}{40}     & 224                                 \\
                                                   & \multicolumn{1}{c|}{Grasp missed} & \multicolumn{1}{c|}{2}      & \multicolumn{1}{c|}{3}       & \multicolumn{1}{c|}{4}      & \multicolumn{1}{c|}{1}       & \multicolumn{1}{c|}{5}      & 15                                  \\
                                                   & \multicolumn{1}{c|}{Collision}    & \multicolumn{1}{c|}{1}      & \multicolumn{1}{c|}{1}       & \multicolumn{1}{c|}{0}      & \multicolumn{1}{c|}{4}       & \multicolumn{1}{c|}{5}      & 11                                  \\ 
                                                   & \multicolumn{1}{c|}{SR} & \multicolumn{1}{c|}{94.0\%} & \multicolumn{1}{c|}{92.0\%}  & \multicolumn{1}{c|}{92.0\%} & \multicolumn{1}{c|}{90.0\%}  & \multicolumn{1}{c|}{80.0\%} & \multicolumn{1}{c}{\textbf{89.6\%}} \\
                                                   & \multicolumn{1}{c|}{SR w.o collision} & \multicolumn{1}{c|}{95.9\%} & \multicolumn{1}{c|}{93.9\%}  & \multicolumn{1}{c|}{92.0\%} & \multicolumn{1}{c|}{97.8\%}  & \multicolumn{1}{c|}{88.9\%} & \multicolumn{1}{c}{\textbf{93.7\%}} \\ \midrule[1.0pt]
\multirow{5}{*}{Low}                               & \multicolumn{1}{c|}{SR}      & \multicolumn{1}{c|}{47}     & \multicolumn{1}{c|}{38}      & \multicolumn{1}{c|}{43}     & \multicolumn{1}{c|}{42}      & \multicolumn{1}{c|}{41}     & 211                                 \\
                                                   & \multicolumn{1}{c|}{Grasp missed} & \multicolumn{1}{c|}{3}      & \multicolumn{1}{c|}{8}       & \multicolumn{1}{c|}{4}      & \multicolumn{1}{c|}{2}       & \multicolumn{1}{c|}{7}      & 24                                  \\
                                                   & \multicolumn{1}{c|}{Collision}    & \multicolumn{1}{c|}{0}      & \multicolumn{1}{c|}{4}       & \multicolumn{1}{c|}{3}      & \multicolumn{1}{c|}{6}       & \multicolumn{1}{c|}{2}      & 15                                  \\ 
                                                   & \multicolumn{1}{c|}{SR} & \multicolumn{1}{c|}{94.0\%} & \multicolumn{1}{c|}{76.0\%}  & \multicolumn{1}{c|}{86.0\%} & \multicolumn{1}{c|}{84.0\%}  & \multicolumn{1}{c|}{82.0\%} & \textbf{84.4\%}  \\
                                                   & \multicolumn{1}{c|}{SR w.o collision} & \multicolumn{1}{c|}{94.0\%} & \multicolumn{1}{c|}{82.6\%}  & \multicolumn{1}{c|}{91.5\%} & \multicolumn{1}{c|}{95.5\%}  & \multicolumn{1}{c|}{85.4\%} & \textbf{89.8\%}                     \\ \midrule[1.0pt]
\multirow{5}{*}{Spot}                              & \multicolumn{1}{c|}{Success}      & \multicolumn{1}{c|}{48}     & \multicolumn{1}{c|}{43}      & \multicolumn{1}{c|}{44}     & \multicolumn{1}{c|}{39}      & \multicolumn{1}{c|}{44}     & 218                                 \\
                                                   & \multicolumn{1}{c|}{Grasp missed} & \multicolumn{1}{c|}{2}      & \multicolumn{1}{c|}{2}       & \multicolumn{1}{c|}{2}      & \multicolumn{1}{c|}{1}       & \multicolumn{1}{c|}{3}      & 10                                  \\
                                                   & \multicolumn{1}{c|}{Collision}    & \multicolumn{1}{c|}{0}      & \multicolumn{1}{c|}{5}       & \multicolumn{1}{c|}{4}      & \multicolumn{1}{c|}{10}      & \multicolumn{1}{c|}{3}      & 22                                  \\  
                                                   & \multicolumn{1}{c|}{SR} & \multicolumn{1}{c|}{96.0\%} & \multicolumn{1}{c|}{86.0\%}  & \multicolumn{1}{c|}{88.0\%} & \multicolumn{1}{c|}{78.0\%}  & \multicolumn{1}{c|}{88.0\%} & \textbf{87.2\%}\\
                                                 & \multicolumn{1}{c|}{SR w.o collision} & \multicolumn{1}{c|}{96.0\%} & \multicolumn{1}{c|}{95.6\%}  & \multicolumn{1}{c|}{95.7\%} & \multicolumn{1}{c|}{97.5\%}  & \multicolumn{1}{c|}{93.6\%} & \textbf{95.6\%} \\   
                                                   
\midrule[1.1pt]
%\multicolumn{1}{l|}{\textbf{Overall success rate}} & \multicolumn{7}{c}{\textbf{87.06\%}}                                                                                                                                                                                            \\ \hline
\multirow{2}{*}{All conditions}                 & \multicolumn{1}{c|}{SR} & \multicolumn{1}{c|}{94.7\%} & \multicolumn{1}{c|}{84.7\%}  & \multicolumn{1}{c|}{88.7\%} & \multicolumn{1}{c|}{84.0\%}  & \multicolumn{1}{c|}{83.3\%} & \textbf{87.06\%}  \\
                                                & \multicolumn{1}{c|}{SR w.o collision} & \multicolumn{1}{c|}{95.3\%} & \multicolumn{1}{c|}{90.7\%}  & \multicolumn{1}{c|}{93.0\%} & \multicolumn{1}{c|}{96.9\%}  & \multicolumn{1}{c|}{89.3\%} & \textbf{93\%}  \\
\bottomrule[1.5pt]

\multicolumn{4}{l}{$^{\mathrm{*}}$ symmetrical object}
\end{tabular}
}}}
\label{grasp-exp}
\end{center}
\end{table*}

In this section, we report the ADD(S) accuracy performance of the proposed two-stage 6D pose estimation algorithm on the LineMOD dataset after training on the synthetic data. We also report the success rate (SR) for grasping different household objects in robotic grasping experiments.

\subsection{6D pose estimation accuracy}

We evaluate the performance of the proposed 6D pose estimation approach on all objects from the LineMOD dataset. We report the results with comparison to the state-of-the-art work in Table \ref{tab: LineMOD_evaluation}, in which the performance of PointFusion is from \cite{lin2022e2ek} and performance of SSD-6D \cite{kehl2017ssd} is from \cite{sundermeyer2018implicit}. Compared to other synthetic-only trained methods, our approach achieves competitive performance with overall 83.6\% pose recognition accuracy without pose refinement. Specially, it performs well on small objects like ``ape", ``duck", on which the SSD-6D \cite{kehl2017ssd} and AAE \cite{sundermeyer2018implicit} are less accurate. On the other hand, our model performs less than optimal on ``holepuncher" and ``camera". The reason could lie in the low-quality textures of the LineMOD models. Our approach, being trained end-to-end on RGBD data, could be more sensitive to less-detailed textures compared to refinement-based approaches. 

Compared to the related work that only uses synthetic data for training, our approach outperforms AAE \cite{sundermeyer2018implicit} and SSD-6D \cite{kehl2017ssd}. Furthermore, \cite{hagelskjaer2021bridging} proposes a 6D pose estimation algorithm based on DGCNN \cite{wang2019dynamic} and reaches 98\% average accuracy. However, it relies heavily on pose refinement, and it takes approximately one second to detect a single object. 

The algorithms trained using real data generally outperform the counterparts trained only on synthetic data, as shown in Table \ref{tab: LineMOD_evaluation}. Nevertheless, we noticed that our approach can achieve approximately 94\% accuracy without refinement when the ground truth bounding box is used to localize the target object, as presented in Table \ref{tab:LineMOD_ablat}. This result is comparable to the state-of-the-art methods trained using real data. One potential way to improve the object detection performance is to use RGBD images \cite{gupta2014learning}, where the object detector can learn the more robust features from both appearances features provided by RGB images and geometry features provided by depth images. This performance gap can also be mitigated by fine-tuning the object detector with a handful of annotated real data. We did not observe a degredation of the object detector on the robotic grasping dataset.

\subsection{Run time}

The efficiency of the proposed approach was evaluated on a workstation equipped with two Xeon Silver-CPU (2.1GHz) and an NVIDIA Quadro RTX 8000 graphics card. 
The results, as reported in Table \ref{time_analysis}, indicate that the inference of PVN3D-tiny consumes the majority of the running time, while the other procedures have similar computational requirements. 
For an input of 480x640 RGB and depth images, the proposed approach has an average running time of 46ms per single object pose estimation. 
This performance is comparable or better than the existing state-of-the-art methods listed in Table \ref{tab: LineMOD_evaluation}, and suitable for real-time robotic tasks. 
As demonstrated in the following section, the accuracy of the proposed approach is sufficient for grasping tasks.

\subsection{6DoF pose estimation in robotic applications}
As shown in Table \ref{grasp-exp}, the robotic arm has achieved an approximate 87\% successful rate (SR). The three scenarios show similar success rates, showing the algorithm's robustness to different lighting levels. Notably, the proposed algorithm works well in low-lighting conditions. The reason could be attributed to two factors: first, the training on domain-randomized synthetic data makes the algorithm learn more robust features. Second, the depth information remains consistent under different lighting conditions, as shown in the Figure \ref{fig:real-grasp}, and so the algorithm can extract sufficient features from depth to compensate for the underexposed color camera.

In general, collision avoidance is not the focus of this research and the results regarding the accuracy of the pose estimation pipeline are more relevant. Thus it is interesting to analyse the grasping success and pose estimation failures excluding all the collision events. The collisions cases are mainly due to the insufficient collision checking. If we neglect the collision cases, the failure cases decrease by 50\%, and the overall grasping performance achieves 93\%. This suggests that a more sophisticated collision checking and grasp pose selection strategy is required and it will be subject of future work.

\textbf{Rubber Duck:} The grasping of the rubber duck is the most robust and successful of all of the objects. The non-regular shape of the duck with no rotational symmetries are robust features, resulting in an accurate pose estimation. Additionally, the rubber material and soft structure facilitate robotic grasping, where slight inaccuracies still lead to successful grasps.

\textbf{Glue bottle:} The glue bottle achieves a good SR as well, due to its shape and material, which are forgiving similarly to that of the rubber duck. Additionally, the bright color of the glue bottle might aid in low light environments, making the object easily visible.

\textbf{Stapler:} The grasping of the stapler is highly affected by the lighting conditions, with 86.0\% SR under spot light and 76.0\% under low light. Possibly, the stapler, due to its dark color, looses more details in low light conditions, making 6D pose estimation more difficult without properly distinguishable features.

\textbf{Chewtoy:} 
During grasping of the chew toy, collision have been the primary cause of failure due to its small size, as it easily gets stuck in small cavities between other objects. We observed this is especially relevant for the chewtoy, because the round shape makes the object roll in the bin, until it gets stopped by other objects. Therefore, the primary reason for failure is not the pose estimation, but the inferior collision avoidance. Moreover under spot lighting conditions, the chewtoy is under-exposed, particularly when stuck in a hole and this makes 6D pose estimation challenging. Combined with the proximity to other objects, this leads to an increased rate of collisions.

\textbf{Pliers:} In the case of grasping the pliers, we observe a higher number of missed grasps, due to the small size of the grasp handles. The grasp generation places all the grasps on the handles and, while according to the ADD most of the proposed grasp would be successful, in practice, some fail in the robotic experiment.
\section{Conclusions and Future Work}
\label{sec:conclusions}

In this work, we introduce 6IMPOSE, a novel framework for sim-to-real data generation and 6D pose estimation. The framework consists of a data generation pipeline that leverages the 3D suite Blender to produce synthetic RGBD image datasets, a real-time two-stage 6D pose estimation approach integrating YOLO-V4-tiny \cite{bochkovskiy2020yolov4} and a real-time version of PVN3D \cite{he2020pvn3d}, and a code base for integration into a robotic grasping experiment.

The results of evaluating the 6IMPOSE framework on the LineMod dataset \cite{hinterstoisser2012model} showed competitive performance with 83.6\% pose recognition accuracy, outperforming or matching state-of-the-art methods. Furthermore, the real-world robotic grasping experiment demonstrated the robustness of the 6IMPOSE framework, achieving an 87\% success rate for grasping five different household objects from cluttered backgrounds under varying lighting conditions.

The contribution of 6IMPOSE lies in its efficient generation of large amounts of photo-realistic RGBD images and successful transfer of the trained inference model to real-world robotic grasping experiments. To the best of our knowledge, this is the first time a sim-to-real 6D pose estimation approach has been systematically and successfully tested in robotic grasping.

In future work, there is potential to further improve 6IMPOSE by exploring improvements to the perception pipeline, such as using a more sophisticated pose detection network or multi-frame detection; improving the scalability and quality of the data generation process; and improving the robotic integration, for example, in areas such as collision detection, grasping pose selection, and support for more scenarios like bin picking where there are multiple instances of the same object class.

\section*{Acknowledgment}
Marco Caccamo was supported by an Alexander von Humboldt Professorship endowed by the German Federal Ministry of Education and Research. 

\bibliographystyle{IEEEtran}
\bibliography{ref}
\vfill
\end{document}